\pdfoutput=1
\documentclass[11pt]{article}
\usepackage[]{acl}
\usepackage{times}
\usepackage{latexsym}
\usepackage[T1]{fontenc}

\usepackage[utf8]{inputenc}
\usepackage{microtype}

\usepackage{booktabs}
\usepackage{multirow}
\usepackage{graphicx}
\usepackage{amsmath}
\usepackage{amssymb}
\usepackage{enumitem}

\title{M$^3$ED: Multi-modal Multi-scene Multi-label \\ Emotional Dialogue Database}

\author{Jinming Zhao$^{1}$, Tenggan Zhang$^{1}$, Jingwen Hu$^{1}$, Yuchen Liu$^{1}$, Qin Jin$^{1}$\footnotemark[1] \\
        {\bf Xinchao Wang$^{3}$, Haizhou Li$^{2,3}$} \\
        $^{1}$ School of Information, Renmin University of China \\ 
        $^{2}$  School of Data Science, The Chinese University of Hong Kong, Shenzhen, China \\ 
        $^{3}$ Electrical and Computer Engineering, National University of Singapore \\
        }

\begin{document}
\maketitle

\renewcommand{\thefootnote}{\fnsymbol{footnote}}
\footnotetext[1]{Corresponding Author}
\renewcommand{\thefootnote}{\arabic{footnote}}

\begin{abstract}
The emotional state of a speaker can be influenced by many different factors in dialogues, such as dialogue scene, dialogue topic, and interlocutor stimulus. The currently available data resources to support such multimodal affective analysis in dialogues are however limited in scale and diversity.
In this work, we propose a \textbf{M}ulti-modal \textbf{M}ulti-scene \textbf{M}ulti-label \textbf{E}motional \textbf{D}ialogue dataset, \textbf{M$^3$ED}, which contains 990 dyadic emotional dialogues from 56 different TV series, a total of 9,082 turns and 24,449 utterances. M$^3$ED is annotated with 7 emotion categories (happy, surprise, sad, disgust, anger, fear, and neutral) at utterance level, and encompasses acoustic, visual, and textual modalities.
To the best of our knowledge, M$^3$ED is the first multimodal emotional dialogue dataset in Chinese.
It is valuable for cross-culture emotion analysis and recognition. 
We apply several state-of-the-art methods on the M$^3$ED dataset to verify the validity and quality of the dataset.
We also propose a general Multimodal Dialogue-aware Interaction framework, MDI, to model the dialogue context for emotion recognition, which achieves comparable performance to the state-of-the-art methods on the M$^3$ED. The full dataset and codes are available\footnote{https://github.com/AIM3-RUC/RUCM3ED}.

\end{abstract}

\section{Introduction}
\begin{figure}[ht]
\centering
\includegraphics[scale=0.55]{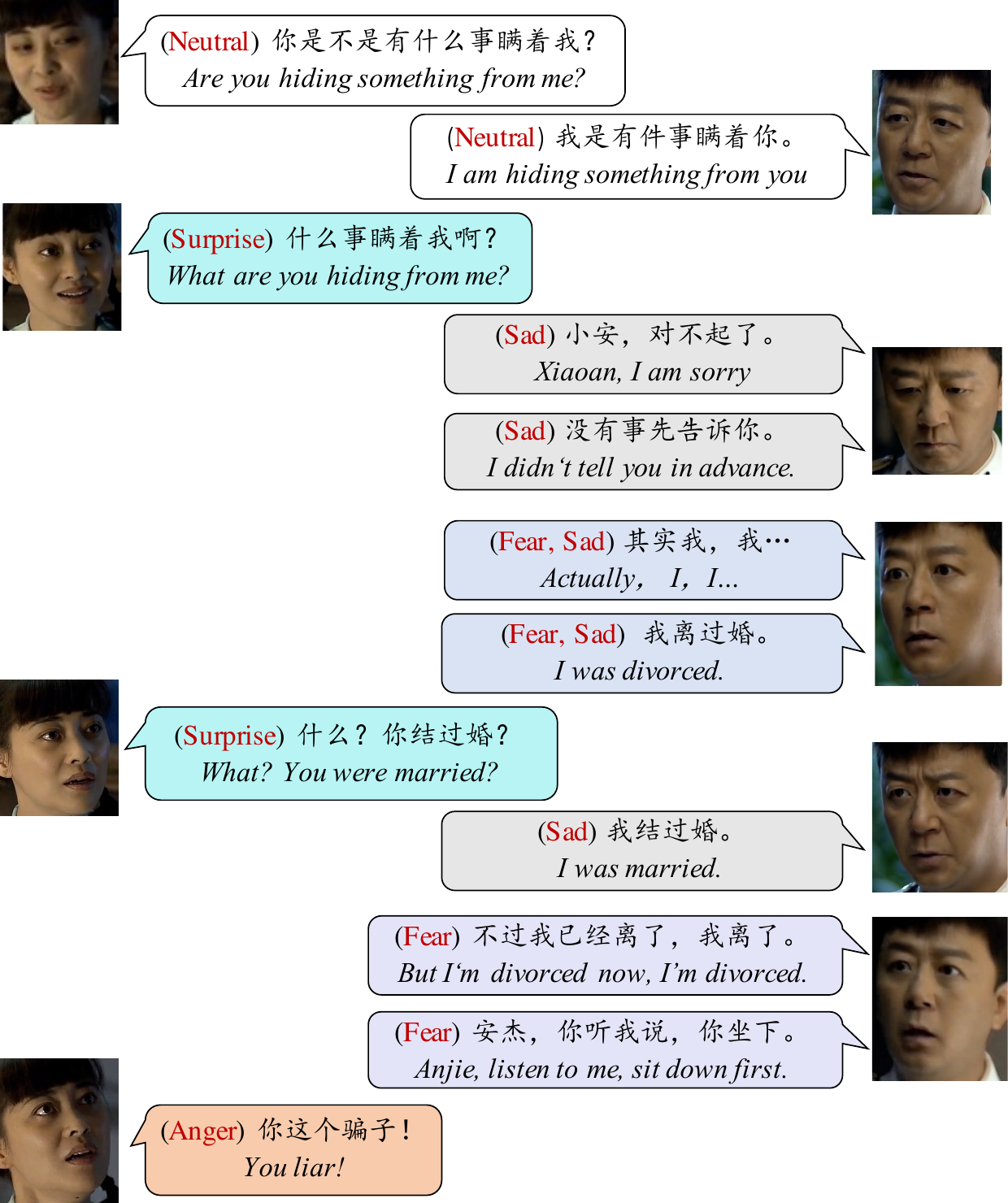}
\caption{\textcolor{black}{An example of a dialogue, showing the rich emotions, inter and intra-turn emotion shifts, emotional inertia and blended emotions.}}
\label{fig:model}
\vspace{-8pt}
\end{figure}

Emotion Recognition in Conversation (ERC) aims to automatically identify and track the emotional status of speakers during a dialogue \cite{poria2019emotion}. It is a crucial component to improve natural human-computer interactions and has a wide range of applications in interaction scenarios, including call-center dialogue systems \cite{danieli2015emotion}, conversational agents \cite{fragopanagos2005emotion} and mental health diagnoses \cite{ringeval2018avec}, etc.
Different from traditional multimodal emotion recognition on isolated utterances, multimodal ERC is a more challenging problem, because there are many influencing factors that affect the speakers' emotional state in a dialogue, including the dialogue context from multi-modalities, the scene, the topic, and even the personality of subjects, etc. \cite{poria2019emotion, scherer2005emotions, koval2015emotional}.
It has been proved in recent works \cite{majumder2019dialoguernn, ghosal2019dialoguegcn, hummgcn, shen2020dialogxl} that contextual information plays an important role in ERC tasks and brings significant improvements over baselines that only consider isolated utterances. 
DialogueRNN \cite{majumder2019dialoguernn} uses recurrent networks to model global and speaker-specific temporal-context information. DialogueGCN \cite{ghosal2019dialoguegcn} and MMGCN \cite{hummgcn} use graph-based networks to capture conversational dependencies between utterances in dialogues. DialogXL \cite{shen2020dialogxl} applies a strong pre-trained language model XLNet \cite{yang2019xlnet} to ERC and proposes a dialog-aware self-attention method for modeling the context information.
The IEMOCAP \cite{busso2008iemocap} and MELD \cite{poria2019meld} are two multimodal emotional dialogue benchmark datasets, which are widely used in the above-mentioned works and promote research in the affective computing field.
However, both of them are limited in size and diversity. The videos in MELD are collected only from the Friends TV series, and the videos in IEMOCAP are recorded in laboratory environments from ten actors performing scripted and spontaneous dialogues. These limitations not only affect the investigation of generalization and robustness of the models, but also limit the exploration of other important influencing factors in dialogues, such as dialogue scene, dialogue topic, emotional influence from interlocutors, and so on.

In this work, we construct a large-scale Multi-modal Multi-scene and Multi-label Emotional Dialogue dataset, \textbf{M$^3$ED}, which consists of 990 emotional dyadic dialogue video clips from 56 different TV series (about 500 episodes), ensuring that there are various dialogue scenes and topics.
We also consider the blended annotations of emotions, which are commonly observed in real-life human interactions \cite{devillers2005challenges, vidrascu2005annotation}.
M$^3$ED contains 24449 utterances in total, which are more than three times larger than IEMOCAP and almost two times larger than MELD.
There are rich emotional interaction phenomena in M$^3$ED dialogues, for example, 5,396 and 2,696 inter-turn emotion-shift and emotion-inertia scenarios respectively, and 2,879 and 10,891 intra-turn emotion-shift and emotion-inertia scenarios respectively.
To the best of our knowledge, M$^3$ED is the first large-scale multi-modal emotional dialogue dataset in Chinese, which can promote research of affective computing for the Chinese language. It is also a valuable addition for cross-cultural emotion analysis and recognition.

We further perform the sanity check of the dataset quality.
Specifically, we evaluate our proposed M$^3$ED dataset on several state-of-the-art approaches, including DialogueRNN, DialogueGCN, and MMGCN. 
The experimental results show that both context information and multiple modalities can help model the speakers' emotional states and significantly improve the recognition performance, in which context information and multiple modalities are two salient factors of a multimodal emotion dialogue dataset.
Furthermore, motivated by the masking strategies of self-attention used in DialogXL \cite{shen2020dialogxl}, we propose a general Multimodal Dialogue-aware Interaction (MDI) framework which considers multimodal fusion, global-local context modeling, and speaker interactions modeling and achieves state-of-the-art performance.

All in all, M$^3$ED is a large, diverse, high-quality, and comprehensive multimodal emotional dialogue dataset, which can support more explorations in the related research directions, such as multi-label learning, interpretability of emotional changes in dialogues, cross-culture emotion recognition, etc. 
The main contributions of this work are as follows:
\begin{itemize}
    \item We build a large-scale Multi-modal Multi-scene and Multi-label Emotional Dialogue dataset called M$^3$ED, which can support more explorations in the affective computing field.
    \item  We perform a comprehensive sanity check of the dataset quality by running several state-of-the-art approaches on M$^3$ED and the experimental results prove the validity and quality of the dataset.
    \item  We propose a general Multimodal Dialogue-aware Interaction framework, MDI, which involves multimodal fusion, global-local context and speaker interaction modeling, and it achieves comparable performance to other state-of-the-art approaches.
\end{itemize}

\section{Related Work}
\label{sec:relatedwork}

\begin{table*}[ht]
\caption{Comparison with existing benchmark datasets. $a, v$ and $l$ refer audio, visual and text respectively.}
\centering
\scalebox{0.78}{
\begin{tabular}{cccccccccc}
\hline

Dataset & Dialogue  & Modalities & Sources & Mul-label & Emos & Spks & Language   & Utts \\ 
\hline
EmoryNLP \cite{zahiri2018emotion} & Yes & $l$ & Friends TV & Yes & 9 & -- & English & 12,606\\
EmotionLines \cite{chen2018emotionlines} & Yes  & $l$ & Friends TV & No & 7 & -- & English  & 29,245 \\
DailyDialog \cite{li2017dailydialog} & Yes  & $l$ & Daily & No & 7 & -- & English  & 102,979 \\
\hline
CMU-MOSEI \cite{zadeh2018mosei} & No & $a,v,l$ & YouTube & No & 7 & 1000 & English & 23,453 \\
AFEW \cite{dhall2012afew} & No & $a,v$ & Movies & No & 7 & 330 & English & 1,645 \\
MEC \cite{li2018mec} & No  & $a,v,l$  & Movies,TVs & No & 8 & -- & Mandarin  & 7,030 \\
CH-SIMS \cite{yu2020ch} & No & $a,v,l$ & Movies,TVs & No & 5 & 474 & Mandarin  & 2,281\\
\hline
IEMOCAP \cite{busso2008iemocap} & Yes & $a,v,l$ & Act & No & 5 & 10 & English  & 7,433 \\
MSP-IMPROV \cite{busso2016msp} & Yes & $a,v,l$ & Act & No & 5 & 12 & English  & 8,438\\
MELD \cite{poria2019meld} & Yes & $a,v,l$ & Friends TV & No & 7 & 407 & English & 13,708 \\
\textbf{M$^3$ED (Ours)} & Yes & $a,v,l$ & 56 TVs & Yes & 7 & 626 & Mandarin & 24,449 \\
\hline
\end{tabular}}
\label{tab:comparison}
\end{table*}

\subsection{Related Datasets}
Table~\ref{tab:comparison} summarizes some of the most important emotion datasets related to this work. 
The EmoryNLP \cite{zahiri2018emotion}, EmotionLines \cite{chen2018emotionlines}, and DailyDialog \cite{li2017dailydialog} are emotional dialogue datasets in only text modality, which have been widely used in the ERC tasks.
The CMU-MOSEI \cite{zadeh2018mosei}, AFEW \cite{dhall2012afew}, MEC \cite{li2018mec}, and CH-SIMS \cite{yu2020ch} contain multiple modalities and have been wildly used for multimodal emotion recognition, but they are not conversational and can not support explorations of dialogue emotional analysis. 
The IEMOCAP \cite{busso2008iemocap}, MSP-IMPROV \cite{busso2016msp} and MELD \cite{poria2019meld} are the currently available multimodal emotional dialogue datasets. 
The IEMOCAP and MSP-IMPROV datasets are recorded from ten/twelve actors performing scripted and spontaneous dyadic dialogues, and each utterance is manually labeled with discrete emotion categories. 
The MELD \cite{poria2019meld} is a multi-modal multi-party emotional dialogue dataset extended from the text-based EmotionLines dataset \cite{chen2018emotionlines}, which is derived only from the Friends TV series.

\subsection{Related Methods}
Previous works on ERC focus on modeling context information in a conversation with different frameworks. BC-LSTM \cite{poria2017context} employs a Bi-directional LSTM to capture temporal-context information in conversations. CMN \cite{hazarika2018conversational} and ICON \cite{hazarika2018icon} use distinct GRUs to model the global and speaker-specific temporal-context, and apply memory networks to model speaker emotional states. DialogueRNN \cite{majumder2019dialoguernn} uses distinct GRUs to model global and speaker-specific temporal-context, and global emotional states tracking respectively.
DialogueGCN \cite{ghosal2019dialoguegcn} captures conversational dependencies between utterances with a graph-based structure. MMGCN \cite{hummgcn} further proposes a GCN-based multimodal fusion method for multimodal ERC tasks to improve recognition performance.
DialogXL \cite{shen2020dialogxl} first introduces a strong pre-trained language model XLNet for text-based ERC. It also proposes several masking strategies of self-attention to model the global, local, inter-speaker, and intra-speaker interactions.
\section{Dataset Construction}
\label{sec:dataConstruction}

\subsection{Dialogue Selection}
In order to build a large-scale, diversified, and high-quality multimodal emotional dialogue dataset, we collect video dialogue clips from different TV series, which can simulate spontaneous emotional behavior in the real-world environment \cite{dhall2012afew, li2018mec, poria2019meld}.

Since high-quality conversation video clips are very important, we require the crowd workers to follow the strict selection requirements, including the following major aspects:
1) The required TV series should belong to these categories, such as family, romance, soap opera, and modern opera, which have rich and natural emotional expressions.
2) The workers are required to select 15 $\sim$ 25 high-quality emotional dialogue video clips from each TV series.
3) Each dialogue should have at least 3 rounds of interaction and a clear conversation topic.
4) In order to ensure the quality of the visual and acoustic modalities, the workers are required to select two-person dialogue scenes with clear facial expressions and intelligible voices.

After the dialogue selection, we randomly check several dialogues for each TV series and filter out the low-quality dialogues or ask the crowd workers to correct the inappropriate start and end timestamps.

\subsection{Annotation}

\subsubsection{Text and Speaker Annotation}
In order to facilitate the process of emotion annotation, we first require the crowd workers to correct the text content and annotate the speaker info of each utterance. Since the videos of TV series do not have embedded subtitles, we use the OCR-based (Optical Character Recognition) method\footnote{The ASR-based methods do not perform as well in these scenarios compared to OCR-based methods.} to automatically generate the text content and the corresponding timestamps. For speaker annotations, the first speaker in the dialogue is annotated as ``A'', and the other speaker is annotated as ``B''. In addition, we annotate the role names, ages and genders of these speakers as well.

\subsubsection{Emotion Annotation}
We annotate each utterance based on Ekman’s six basic emotions (\textit{happy, surprise, sad, disgust, anger, and fear}) and an additional emotion label \textit{neutral}, which is an annotation scheme widely used in previous works \cite{poria2019meld, busso2008iemocap}.
The annotators are asked to sequentially annotate the utterances, after watching the videos. Thus, the textual, acoustic and visual information, and the previous utterances in the dialogue are available for emotional annotation.
The annotators are allowed to select more than one emotional label to account for blended emotions (e.g., anger\&sad), which are commonly observed in real-life human interactions \cite{devillers2005challenges}.
If none of the seven emotion categories can accurately describe the emotion status of the utterance, a special $other$ category can be annotated.

In order to obtain high-quality annotations, we together with several emotional psychology experts design an annotation tutorial with reference to previous guidelines~\cite{ekman1992argument, campos2013shared}. We train the annotators and provide them with an examination, and only those who pass the exam can participate in the annotation stage. The vast majority of the dataset is annotated by university students and all the annotators are native Mandarin speakers. We assign three annotators to each dialogue.

\begin{table}[t]
\caption{M$^3$ED statistics. ``x-turn'' and ``in-turn'' refer to \textit{inter-turn} and \textit{intra-turn} respectively.}
\vspace{-6pt}
\centering
\scalebox{0.83}{
\begin{tabular}{cccc|c}
\hline
Statistics & Train  & Val  & Test & Total \\ 
\hline
\# TV series & 38 & 7 & 11 & 56 \\
\# dialogs & 685 & 126 & 179 & 990 \\
\# turns & 6,505 & 1,016 & 1,561 & 9,082 \\
\# utts & 17,427 & 2,821 & 4,201 &  24,449\\
\# spkrs & 421 & 87 & 118 & 626 \\
Avg. turns/dialog & 9.5 & 8.06 & 8.72 & 9.17 \\
Avg. utts/turns & 2.68 & 2.78 & 2.69 & 2.69 \\
Avg. utts/dialog & 25.44 & 22.39 & 23.47 & 24.7 \\
Avg. utt length & 7.41 & 7.28 & 7.42 & 7.39\\
Avg. dur/dialog & 53.22 & 46.68 & 47.94 & 51.43 \\
\# x-turn emo-shift & 3,854 & 622 & 920 & 5,396 \\
\# x-turn emo-inertia & 1,966 & 268 & 462 & 2,696 \\
\# in-turn emo-shift & 2,029 & 338  & 512 & 2,879 \\
\# in-turn emo-inertia & 7,775 & 1,292 & 1,824 & 10,891 \\
\# blended emos  & 1,862 & 379 & 386 & 2,627 \\
Fleiss' Kappa & 0.59 & 0.603 & 0.579 & 0.59 \\
\hline
\end{tabular}}
\label{tab:statistics}
\end{table}
\vspace{-7pt}

\begin{table}[h]
\caption{Emotion Distribution of M$^3$ED.}
\vspace{-6pt}
\centering
\scalebox{0.9}{
\begin{tabular}{ccccc}
\hline
Emotion & Train  & Val  & Test & Total \\
\hline
neutral &7,130  & 1,043 & 1,855 & 10,028 \\
happy & 1,626 & 303 & 358 & 2,287 \\
surprise & 696 & 120 & 235 & 1,051 \\
sad & 2,734 & 489 & 734 & 3,957 \\
disgust & 1,145 & 134 & 218 & 1,497 \\
anger & 3,816 & 682 & 736 & 5,234 \\
fear & 280 & 50 & 65 & 395 \\
\hline
Total & 17,427 & 2,821 & 4,201 & 24,449 \\
\hline
\end{tabular}}
\label{tab:emo_distribution}
\end{table}

\begin{table}[h]
\caption{Speaker/Age/Gender Distributions of M$^3$ED.}
\vspace{-6pt}
\scalebox{0.9}{
\begin{tabular}{c|c}
\hline
Speakers & 626 \\
Gender & Male: 328 Female: 298 \\
Age & Child: 34 Young: 295 Mid: 223 Old: 74 \\
\hline
\end{tabular}}
\label{tab:age_gender_distribution}
\end{table}

\subsection{Emotion Annotation Finalization}

We apply the majority voting strategy over all the annotations of an utterance to produce its final emotion label. Please note that annotators are allowed to assign more than one emotion label to an utterance, and the importance of these labels is in descending order. We simply assign an importance value to the emotion label of each utterance in descending order, e.g. $I(e)=7$ for the first emotion label, $I(e)=6$ for the second emotion label, and so on. If a label is not assigned to the utterance, its importance value $I(e)=0$.
An emotion label $e$ is assigned as one of the final emotion labels for an utterance, if it is assigned to the utterance by at least two annotators. And its importance value is decided by averaging its importance ranking from all annotators: $I(e) = \sum_{k=1}^3 I_k(e)$, where $I_k(e)$ is its importance value from annotator $k$.

To further ensure annotation quality, we design two strategies to review and revise incorrect annotations. 1) We calculate the annotation agreement between the annotators of each dialogue. For the dialogues with a poor agreement, we require all relevant annotators to review the annotations again and make corrections if necessary. 2) For the utterances (0.5\% of all utterances) that don't have a majority annotators' agreement, we ask several high-quality annotators to review them and make a final emotion annotation decision for these utterances.

\begin{figure*}[t]
\centering
\includegraphics[scale=0.8]{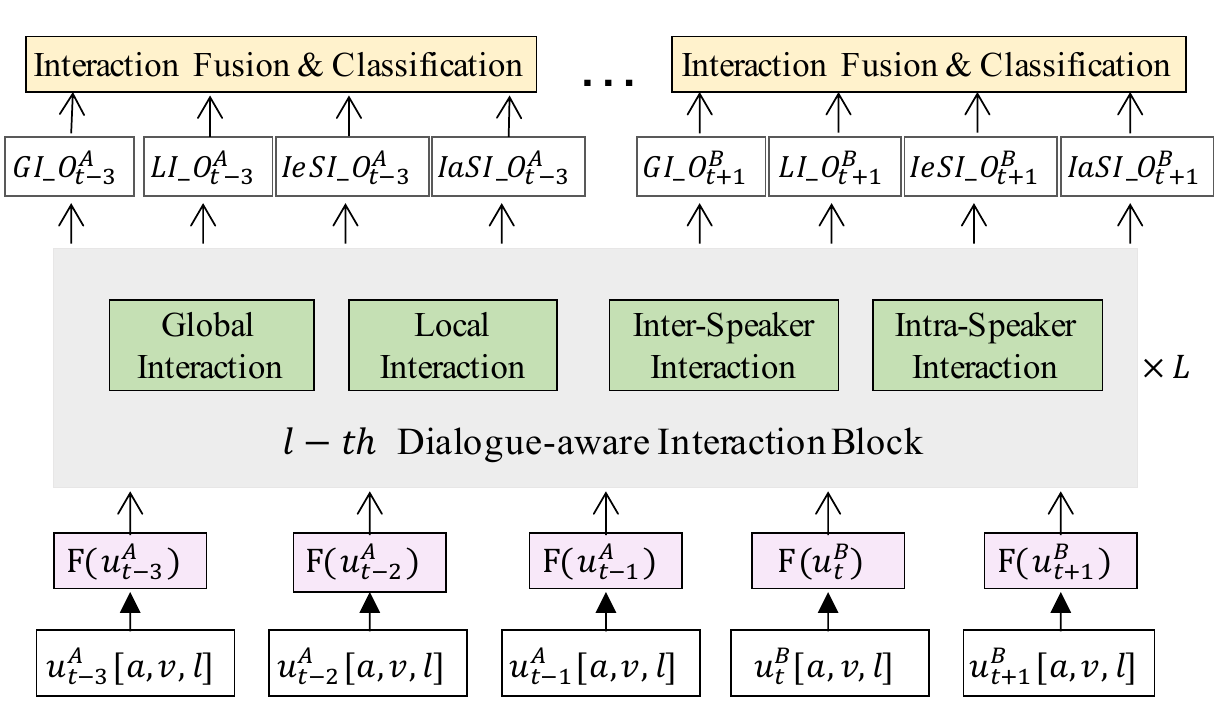}
\caption{Illustration of the Multimodal Dialog-aware Interaction (MDI) framework (taking one round as an example). $l$ represents the $l$-th block in the Dialog-aware Interaction Module. $F(\cdot)$ denotes the multimodal fusion module. The $GI\_O^{B}_{t}$ represents the output of the $t$-th utterance from the Global Interaction of Dialog-aware Interaction Module. Similarly, the $LI\_O$, $IeSI\_O$, and $IaSI\_O$ represent the output of the Local Interaction, Inter-Speaker Interaction and Intra-Speaker Interaction respectively.}
\label{fig:model}
\vspace{-5pt}
\end{figure*}

Finally, we analyze the inter-annotators agreement and achieve an overall Fleiss’ Kappa \cite{fleiss2013statistical} statistic of $k=0.59$ for a seven-class emotion problem, which is higher than other datasets, such as $k=0.43$ in MELD, $k=0.48$ in IEMOCAP and $k=0.49$ in MSP-IMPROV.

\subsection{Dataset Statistics}
\label{data:statis}
Table~\ref{tab:statistics} presents several basic statistics of the M$^3$ED dataset. 
It contains 990 dialogues, 9,082 turns, 24,449 utterances derived from 56 different TV series (about 500 episodes), which ensures the scale and diversity of the dataset.
We adopt the TV-independent data split manner in order to avoid any TV-dependent bias, which means there is no overlap of TV series across training, validation, and testing sets. The basic statistics are similar across these three data splits.
There are rich emotional interactions phenomena in the M$^3$ED, for example, 5,396 and 2,696 inter-turn emotion-shift and emotion-inertia scenarios respectively, and 2,879 and 10,891 intra-turn emotion-shift and emotion-inertia scenarios. The emotion shift and emotion inertia are two important factors in dialogues, which are challenging and worthy of exploration \cite{poria2019meld}.
As shown in the table, 89\% of utterances are assigned with one emotion label, and 11\% of utterances are assigned with blended emotions\footnote{The top 5 most frequent blended emotions are: anger\&disgust, anger\&sad, sad\&anger, disgust\&anger and fear\&sad}.

Table~\ref{tab:emo_distribution} presents the single emotion distribution statistics.
The distribution of each emotion category is similar across train/val/test sets.
As shown in Table~\ref{tab:age_gender_distribution}, there are in total 626 different speakers in M$^3$ED with balanced gender distribution. Among all the speakers,  young and middle-aged speakers account for more than 80\%.

\section{Proposed Framework}
\label{sec:method}
A dialogue can be defined as a sequence of utterances $D=\{utt_{1}, utt_{2},..., utt_{N}\}$, where $N$ is the number of utterances. Each utterance consists of textual ($l$), acoustic ($a$) and visual ($v$) modalities.
We denote $u^{A}_t[a,v,l]$ as the utterance-level feature of utterance $utt_{t}$ from speaker A with the textual, acoustic and visual modality respectively.
The task aims to predict the emotional state for each utterance in the dialogue based on all existing modalities. 
Figure~\ref{fig:model} illustrates our proposed Multimodal Dialogue-aware Interaction (MDI) framework, which contains three main modules: 1) \textit{Multimodal Fusion} module aims to generate the utterance-level multimodal representation from different modalities. 2) \textit{Dialog-aware Interaction} module aims to model the interactions in the dialogue; 3) \textit{Interaction Fusion and Classification} module fuses the different interaction information from the outputs of the Dialog-aware Interaction module, and then makes the emotional state prediction based on the fused interaction information.

\noindent \textbf{Multimodal Fusion Module:}
Based on the modality-specific feature representations from different modalities, we apply early fusion of these modalities features to produce the multimodal feature representation: $u = \text{concat}(u[a], u[v], u[l])$.

\begin{figure}[t]
    \centering
    \begin{minipage}{0.5\linewidth}
      \centerline{\includegraphics[width=2.8cm]{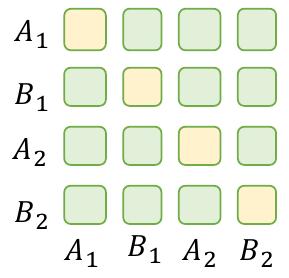}}
      \centerline{\small{(a) Global Mask}}
    \end{minipage}
    \begin{minipage}{0.45\linewidth}
      \centerline{\includegraphics[width=2.8cm]{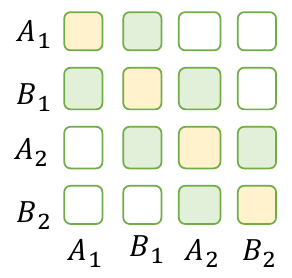}}
      \centerline{\small{(b) Local Mask (window=1)}}
    \end{minipage}
    \\
    \centering
      \begin{minipage}{0.5\linewidth}
      \centerline{\includegraphics[width=2.8cm]{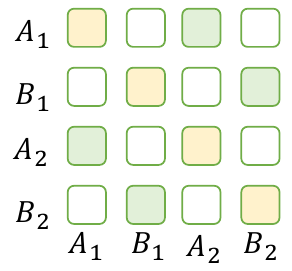}}
      \centerline{\small{(c) Intra-speaker Mask}}
    \end{minipage}
    \begin{minipage}{0.45\linewidth}
      \centerline{\includegraphics[width=2.8cm]{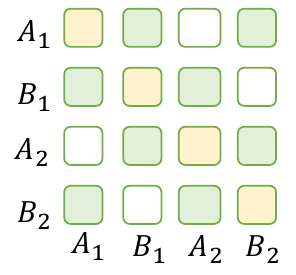}}
      \centerline{\small{(d) Inter-speaker Mask}}
    \end{minipage}
    \caption{Illustration of the four masking strategies corresponding to the four interaction sub-modules respectively. $A_{i}$ denotes the $i$-th utterance of the speaker A. The yellow blocks denote the current utterances. Utterances that can be accessed by the current utterance are marked as green, while those can not be accessed are marked as white.}
    \label{fig:interact}
    \vspace{-3pt}
\end{figure}

\noindent \textbf{Dialog-aware Interaction Module:}
In order to adequately capture the contextual information in the dialogue, we propose the Dialog-aware Interaction Module which consists of $L$ dialog-aware interaction blocks (gray block in Figure~\ref{fig:model}).
In each block, we adopt four sub-modules, Global Interaction, Local Interaction, Intra-speaker Interaction and Inter-speaker Interaction, to model the global, local, intra-speaker and inter-speaker interactions in the dialogue respectively. We implement these four types of interactions in one Transformer layer by skillfully changing the masking strategies of self-attention \cite{shen2020dialogxl, li2020hierarchical} as illustrated in Figure~\ref{fig:interact}.

\vspace{10pt}
\noindent \textbf{Interaction Fusion and Classification:}
As the Dialog-aware Interaction Module produces different outputs that carry various interaction contextual information, we fuse these outputs via simple addition. Finally, we use one fully connected layer as a classifier to predict the emotional state based on the fused interaction information.
\section{Experiments}

\subsection{Feature Extraction}
\label{sec:expr:feature}
We investigate the state-of-the-art features of different modalities including textual, acoustic, and visual features for emotion recognition tasks\footnote{More detailed description of the feature extractors can be found in the supplementary material.~\ref{sup:features}}.

\textbf{Textual Features:} We extract the word-level features from a pre-trained RoBERTa model \cite{yu2020ch}. Furthermore, to get more efficient emotional features, we extract the finetuned features (``[CLS]'' position) from the finetuned RoBERTa model trained on M$^3$ED. We refer to the word-level and finetuned utterance-level textual features as ``L\_Frm'', and ``L\_Utt'' respectively.

\textbf{Acoustic Features:} We extract the frame-level features from a pre-trained Wav2Vec2.0 model \cite{baevski2020wav2vec}. We extract the finetuned features (the last time step) from the  Wav2Vec2.0 model finetuned on M$^3$ED. We refer to the frame-level and finetuned utterance-level acoustic features as ``A\_Frm'' and ``A\_Utt'' respectively.

\textbf{Visual Features:} We first propose a two-stage strategy to detect the speaker’s faces\footnote{more details in supplementary material.~\ref{sup:asd}}. We then extract the face-level features via a pre-trained DenseNet model \cite{huang2017densely} for each utterance based on the detected speaker's faces. DenseNet was trained on two facial expression benchmark corpus, FER+ \cite{barsoum2016training} and AffectNet \cite{mollahosseini2017affectnet}.
We average the face-level features within one utterance to get the averaged utterance-level features. We refer to the face-level, averaged utterance-level visual features as ``V\_Frm'', ``V\_Utt'' respectively.

\subsection{Baseline Models}
\label{sec:exprs:baselines}
We evaluate several state-of-the-art methods including utterance-level recognition methods and dialog-level recognition methods on the proposed M$^3$ED dataset, and they are listed as follows:

\textbf{MultiEnc:} A flexible and efficient utterance-level multimodal emotion recognition framework \cite{zhaomissing} that consists of several modality-specific encoders (LSTM, LSTM and TextCNN for acoustic, visual and textual modalities respectively) and a fusion encoder (several fully-connected layers) for emotion prediction. For the utterance-level modality features, three DNN encoders are used for the three modalities respectively.

\textbf{DialogueRNN:} A state-of-the-art RNN-based ERC framework proposed in \cite{majumder2019dialoguernn}, which captures the global and speaker-specific temporal context information, and global emotional state information via different GRUs. For the multimodal experiments, the early-fusion method that concatenates different modality features as input is adopted in this work.

\textbf{DialogueGCN:} A state-of-the-art GCN-based ERC framework proposed in \cite{ghosal2019dialoguegcn}, which models long-distance dependency and speaker interactions via direct edges and different designed relations respectively. For the multimodal experiments, we also adopt the early-fusion method in this work.

\textbf{MMGCN:} A state-of-the-art GCN-based multimodal ERC framework proposed in \cite{hummgcn}. For the uni-modal experiments, we only model the fully connected graph.

\vspace{-3pt}
\subsection{Experiment Setup}
We split the M$^3$ED dataset into training, validation, testing sets in a TV-independent manner, which is a more challenging experiment setting. The distribution of the data splits is shown in Table~\ref{tab:emo_distribution}. We use the weighted-F1 score (WF1) as the evaluation metrics. We tune the parameters on the validation set and report the performance on the testing set. We run each model three times and report the average performance to alleviate the influence of random parameter initialization.

We conduct two sets of experiments, including 1) the utterance-level baseline experiments of emotion recognition on isolated utterances without considering dialogue context, which aims to check the quality of each modality and compare the effectiveness of multimodal information for emotion recognition, 
and 2) the dialogue-level experiments of emotion recognition in the dialogue, which aims to compare our proposed general MDI framework with the state-of-the-art models in modeling dialogue context for emotion recognition.
For the utterance-level experiments, we adopt the MultiEnc (Section~\ref{sec:exprs:baselines}) framework as the baseline model. For the dialogue-level experiments, we compare to DialogueRNN, DialogueGCN, and MMGCN models.

Since different modality features are used in this work, we have tried different hidden sizes (such as 180, 256, and 512) in our experiments. For the experiments on the proposed Multimodal Dialog-aware Interaction framework (Section~\ref{sec:method}), we use the Adam optimizer with learning rate of 3e-5. We set the dropout as 0.1, the hidden size as 384 in the unimodal experiments and 512 in the multimodal experiments.

\subsection{Utterance Baseline Experiments}
\label{sec:exprs:utt}
Table~\ref{tab:utt_baseline_exprs} presents the utterance-level baseline results. 
Among the different unimodal features, the finetuned utterance-level features achieve significant improvement on textual and acoustic modalities. The multimodal information can bring significant performance improvement over unimodal. However, for the multimodal experiments, the finetuned features do not show much improvement over the frame-level features. It is mainly because the finetuned features retain more classification information and lose some modality-specific information, which limits the complementarity between the modalities.

In addition, we observe that there is no big gap between the performances on different modalities, which indicates the good quality of different modalities in our M$^3$ED dataset.

\begin{table}[t]
\caption{Utterance-level baseline performance (WF1) of different features and different modalities. ``Frm'', ``Utt'' refer to frame-level, utterance-level features respectively.}
\centering
\scalebox{0.8}{
\begin{tabular}{c|cc|cc}
\hline
Modalities & \multicolumn{2}{c|}{Frm}  & \multicolumn{2}{c}{Utt}  \\ 
     & val    & test   & val & test  \\
\hline
$\{l\}$   & 42.24  & 43.23  &  44.67 & \textbf{44.41} \\
$\{a\}$   & 42.56  & 40.96  &  48.56 & \textbf{46.09} \\
$\{v\}$   & 43.79  & \textbf{41.25}  &  42.32 & 41.09 \\
\hline
$\{l,a\}$   & 48.10  & 46.53  & 51.58 & \textbf{48.68}  \\
$\{l,v\}$  & 50.73  & \textbf{48.17}   &  50.48 & 47.68  \\
$\{a,v\}$   & 49.66  & 46.19   &  49.66 & \textbf{46.28}  \\
\hline
$\{l,a,v\}$  & 54.55  & \textbf{49.48} & 52.15 & 48.90 \\
\hline
\end{tabular}
\label{tab:utt_baseline_exprs}
}
\end{table}

\begin{table*}[ht]
    \caption{Emotion recognition performance (WF1) in dialogues under the unimodal and multimodal conditions.}
    \centering
\scalebox{0.78}{
    \begin{tabular}{|c|c|ccccccc|}
    \hline
    \multirow{2}{*}{Model} & \multirow{2}{*}{Metric} & \multicolumn{7}{c|}{Modalities} \\ \cline{3-9} 
                                         & & $\{l\}$ & $\{a\}$  & $\{v\}$ & $\{l,a\}$ & $\{l,v\}$   & $\{a,v\}$  & $\{l,a,v\}$ \\
     \hline
    \multirow{2}{*}{UttBaseline}& val     & 44.67   &  48.56  &  42.32  &  51.58     &  50.48  & 49.66   &  52.15  \\
                                 & test    & 44.41   &  46.09  &  41.09  &  48.68     &  47.68  & 46.28  &   48.90  \\
    \hline
    \multirow{2}{*}{DialogueGCN} & val     & 50.77 (\textcolor{blue}{+6.1}) &  50.96 (\textcolor{blue}{+2.4})  &  33.82  &  53.91 (\textcolor{blue}{+2.3})  &  54.26 (\textcolor{blue}{+3.8})   &  50.80 (\textcolor{blue}{+1.1})  &  54.58 (\textcolor{blue}{+2.4}) \\
                                 & test    & 46.09 (\textcolor{blue}{+1.7}) &  46.45 (\textcolor{blue}{+0.4})  &  27.79  &  49.44 (\textcolor{blue}{+0.8})  &  49.26 (\textcolor{blue}{+1.6})   &  47.09 (\textcolor{blue}{+0.8}) &  49.93 (\textcolor{blue}{+1.0}) \\
    \hline
    \multirow{2}{*}{MMGCN} & val    & 50.83 (\textcolor{blue}{+6.2}) &  52.93 (\textcolor{blue}{+4.4})  &  37.05  &  55.62 (\textcolor{blue}{+4.0})   &  54.75 (\textcolor{blue}{+4.3}) &  54.71 (\textcolor{blue}{+5.1})  &  56.67 (\textcolor{blue}{+4.5}) \\
                          & test    & 46.49 (\textcolor{blue}{+2.1}) &  47.77 (\textcolor{blue}{+1.7})  &  32.87  &  49.44 (\textcolor{blue}{+0.8})   &  50.42 (\textcolor{blue}{+2.7})   &  \textbf{48.55} (\textcolor{blue}{+2.3})  &  51.18 (\textcolor{blue}{+2.3}) \\
    \hline
    \multirow{2}{*}{DialogueRNN} & val     & 53.65 (\textcolor{blue}{+9.0}) &  52.09 (\textcolor{blue}{+3.5}) &  36.03  &  55.85 (\textcolor{blue}{+4.3})  &  58.70 (\textcolor{blue}{+8.2})  &  52.69 (\textcolor{blue}{+3.0})  &  56.52 (\textcolor{blue}{+4.4})  \\
                                 & test    & 48.80 (\textcolor{blue}{+4.4}) &  47.65 (\textcolor{blue}{+1.6}) &  28.38  &  \textbf{51.87} (\textcolor{blue}{+3.2})  & \textbf{52.28} (\textcolor{blue}{+4.6})   &  47.49 (\textcolor{blue}{+1.2})  &  \textbf{51.66} (\textcolor{blue}{+2.8}) \\
    \hline
    \multirow{2}{*}{Ours} & val     & 51.37 (\textcolor{blue}{+6.7}) & 51.5 (\textcolor{blue}{+2.9})  &  45.97 (\textcolor{blue}{+3.6})  &  54.27 (\textcolor{blue}{+2.7}) &  55.69 (\textcolor{blue}{+5.2}) &  51.34 (\textcolor{blue}{+1.7}) &  54.09 (\textcolor{blue}{+1.9})  \\
                          & test    & \textbf{49.42} (\textcolor{blue}{+5.0}) &  \textbf{48.03} (\textcolor{blue}{+1.9})  &  \textbf{41.33} (\textcolor{blue}{+0.2})  &  50.24 (\textcolor{blue}{+1.6}) &  52.07 (\textcolor{blue}{+4.4}) &  47.64 (\textcolor{blue}{+1.4}) &  50.99 (\textcolor{blue}{+2.1})  \\
    \hline
    \end{tabular}
    \label{tab:dialog_exprs}
}
\end{table*}

\subsection{Dialogue Experiments}

Since the state-of-the-art dialogue-level methods mainly focus on modeling the dialogue context information based on the utterance-level features, we adopt the finetuned utterance-level features (``Utt\_ft'') in the following experiments. Table~\ref{tab:dialog_exprs} presents the dialogue-level experiment results. The results show that context information and multiple modalities, the two salient factors of a multimodal emotion dialogue dataset, both bring significant performance improvement, which also proves the validity and quality of the M$^3$ED dataset to some extend.

Compared to the state-of-the-art models, our proposed general MDI framework achieves superior performance in the textual, acoustic, and visual unimodal experiments. It demonstrates that the four dialogue-aware interaction strategies which consider both the global- and local-context interactions and the intra- and inter-speaker interactions have better dialogue modeling ability than only considering part of these interactions, which demonstrates the strong dialogue context modeling ability of MDI.
However, MDI does not outperform other models under the multimodal conditions, which may be due to the limited training dataset size and the limited ability of the vanilla multimodal fusion strategy in interaction modeling. In the future, we will explore more effective multimodal fusion module and interaction modeling module within the MDI framework to improve its performance under multimodal conditions.

\section{Future Directions}
\label{sec:future}

The M$^3$ED dataset is a large, diversified, high-quality, and comprehensive multimodal emotional dialogue dataset. Based on the characteristics of the dataset and the analysis from the extensive experiments, we believe that M$^3$ED can support a number of related explorations in affective computing field.

\begin{itemize}[leftmargin=*]
    \setlength{\itemsep}{0pt}
    \item Based on the experiment results, we think that the finetuned features lack sufficient modality-specific information, which limits the performance under the multimodal conditions. Therefore, it is worth exploring to realize a more efficient multimodal fusion module based on the raw frame-level features and make the above proposed general Multimodal Dialog-aware Interaction (MDI) framework an end-to-end model. 
    \item According to psychological and behavioral studies, emotional inertia and stimulus (external/internal) are important factors that affect the speaker’s emotional state in dialogues. The emotional inertia and emotional stimulus can explain how one speaker's emotion affects his own or the other speaker's emotion. There are rich emotional interaction phenomena including inter- and intra-turn emotion shifts in the M$^3$ED dataset. Therefore, it can support the exploration of interpretability of emotional changes in a Dialogue.
    \item The blended emotions are commonly observed in human real-life dialogues, and multi-label learning can help reveal and model the relevance between different emotions. Therefore, the M$^3$ED dataset can support the exploration of multi-label emotion recognition in conversations.
    \item Emotional expression varies across different languages and cultures. The M$^3$ED dataset in Chinese is a valuable addition to the existing benchmark datasets in other languages. It can promote the research of cross-culture emotion analysis and recognition.

\end{itemize}
\section{Conclusion}
\label{sec:conclusion}
In this work, we propose a multi-modal, multi-scene, and multi-label emotional dialogue dataset, M$^3$ED, for multimodal emotion recognition in conversations.
Compared to MELD, the currently largest multimodal dialogue dataset for emotion recognition, M$^3$ED is larger (24,449 vs. 13,708 utterances), more diversified (56 different TV series vs. only one TV series Friends), with higher-quality (balanced performance across all three modalities), and containing blended emotions annotation which is not available in MELD. 
M$^3$ED is the first multi-modal emotion dialogue dataset in Chinese, which can serve as a valuable addition to the affective computing community and promote the research of cross-culture emotion analysis and recognition.
Furthermore, we propose a general Multimodal Dialog-aware Interaction framework, which considers multimodal fusion, temporal-context modeling, and speaker interactions modeling, and achieves the state-of-the-art performance.
We also propose several interesting future exploration directions based on the M$^3$ED dataset.

\section{Acknowledgments}
This work was partially supported by the National Key R\&D Program of China (No. 2020AAA0108600), the National Natural Science Foundation of China (No. 62072462), Large-Scale Pre-Training Program 468 of Beijing Academy of Artificial Intelligence (BAAI), A*STAR RIE2020 Advanced Manufacturing and Engineering Domain (AME) Programmatic Grant (No. A1687b0033), NRF Centre for Advanced Robotics Technology Innovation (CARTIN) Project and China Scholarship Council.

\section{Ethical Considerations}
This work presents M$^3$ED, free and open dataset for the research community to study the multimodal emotion recognition in dialogues.
Data in  M$^3$ED are collected from TV series in Chinese.
To ensure that crowd workers were fairly compensated, we paid them at an hourly rate of 40 yuan (\$6.25 USD) per hour, which is a fair and reasonable hourly wage in Beijing.
First, to select high-quality dialogues from 56 TV-series, we recruited 12 Chinese college students (5 males and 7 females).
Each student was paid 100 yuan (\$15.625 USD) for selecting about 18 dialogues from each TV series.
To annotate the emotional status of the selected dialogues, we recruited 14 Chinese college students (6 males and 8 females).
Each student was paid 200 yuan (\$31.25 USD) for annotating about 18 dialogues from each TV series with emotion labels, text correction, speaker, gender, and age information. If only the emotion labels were annotated, the payment for each TV series was 100 yuan (\$15.625 USD).
Considering the copy-right issue of TV-series, we will only release the name list of the TV-series and our annotations. To facilitate future comparison research on this dataset, we will provide our extracted visual expression features and acoustic features. We anticipate that the high-quality and rich annotation labels in the dataset will advance research in multimodal emotion recognition.


\bibliography{anthology,custom}

\begin{thebibliography}{36}
\expandafter\ifx\csname natexlab\endcsname\relax\def\natexlab#1{#1}\fi

\bibitem[{Baevski et~al.(2020)Baevski, Zhou, Mohamed, and
  Auli}]{baevski2020wav2vec}
Alexei Baevski, Henry Zhou, Abdelrahman Mohamed, and Michael Auli. 2020.
\newblock wav2vec 2.0: A framework for self-supervised learning of speech
  representations.
\newblock \emph{arXiv preprint arXiv:2006.11477}.

\bibitem[{Barsoum et~al.(2016)Barsoum, Zhang, Ferrer, and
  Zhang}]{barsoum2016training}
Emad Barsoum, Cha Zhang, Cristian~Canton Ferrer, and Zhengyou Zhang. 2016.
\newblock Training deep networks for facial expression recognition with
  crowd-sourced label distribution.
\newblock In \emph{Proceedings of the 18th ACM International Conference on
  Multimodal Interaction}, pages 279--283.

\bibitem[{Busso et~al.(2008)Busso, Bulut, Lee, Kazemzadeh, Mower, Kim, Chang,
  Lee, and Narayanan}]{busso2008iemocap}
Carlos Busso, Murtaza Bulut, Chi-Chun Lee, Abe Kazemzadeh, Emily Mower, Samuel
  Kim, Jeannette~N Chang, Sungbok Lee, and Shrikanth~S Narayanan. 2008.
\newblock Iemocap: Interactive emotional dyadic motion capture database.
\newblock \emph{Language resources and evaluation}, 42(4):335--359.

\bibitem[{Busso et~al.(2016)Busso, Parthasarathy, Burmania, AbdelWahab,
  Sadoughi, and Provost}]{busso2016msp}
Carlos Busso, Srinivas Parthasarathy, Alec Burmania, Mohammed AbdelWahab,
  Najmeh Sadoughi, and Emily~Mower Provost. 2016.
\newblock Msp-improv: An acted corpus of dyadic interactions to study emotion
  perception.
\newblock \emph{IEEE Transactions on Affective Computing}, 8(1):67--80.

\bibitem[{Campos et~al.(2013)Campos, Shiota, Keltner, Gonzaga, and
  Goetz}]{campos2013shared}
Belinda Campos, Michelle~N Shiota, Dacher Keltner, Gian~C Gonzaga, and
  Jennifer~L Goetz. 2013.
\newblock What is shared, what is different? core relational themes and
  expressive displays of eight positive emotions.
\newblock \emph{Cognition \& emotion}, 27(1):37--52.

\bibitem[{Chen et~al.(2018)Chen, Hsu, Kuo, Ku et~al.}]{chen2018emotionlines}
Sheng-Yeh Chen, Chao-Chun Hsu, Chuan-Chun Kuo, Lun-Wei Ku, et~al. 2018.
\newblock Emotionlines: An emotion corpus of multi-party conversations.
\newblock \emph{arXiv preprint arXiv:1802.08379}.

\bibitem[{Danieli et~al.(2015)Danieli, Riccardi, and Alam}]{danieli2015emotion}
Morena Danieli, Giuseppe Riccardi, and Firoj Alam. 2015.
\newblock Emotion unfolding and affective scenes: A case study in spoken
  conversations.
\newblock In \emph{Proceedings of the International Workshop on Emotion
  Representations and Modelling for Companion Technologies}, pages 5--11.

\bibitem[{Devillers et~al.(2005)Devillers, Vidrascu, and
  Lamel}]{devillers2005challenges}
Laurence Devillers, Laurence Vidrascu, and Lori Lamel. 2005.
\newblock Challenges in real-life emotion annotation and machine learning based
  detection.
\newblock \emph{Neural Networks}, 18(4):407--422.

\bibitem[{Dhall et~al.(2012)Dhall, Goecke, Lucey, and Gedeon}]{dhall2012afew}
Abhinav Dhall, Roland Goecke, Simon Lucey, and Tom Gedeon. 2012.
\newblock Collecting large, richly annotated facial-expression databases from
  movies.
\newblock \emph{IEEE multimedia}, 19(03):34--41.

\bibitem[{Ekman(1992)}]{ekman1992argument}
Paul Ekman. 1992.
\newblock An argument for basic emotions.
\newblock \emph{Cognition \& emotion}, 6(3-4):169--200.

\bibitem[{Fleiss et~al.(2013)Fleiss, Levin, and Paik}]{fleiss2013statistical}
Joseph~L Fleiss, Bruce Levin, and Myunghee~Cho Paik. 2013.
\newblock \emph{Statistical methods for rates and proportions}.
\newblock john wiley \& sons.

\bibitem[{Fragopanagos and Taylor(2005)}]{fragopanagos2005emotion}
Nickolaos Fragopanagos and John~G Taylor. 2005.
\newblock Emotion recognition in human-computer interaction.
\newblock \emph{Neural Networks}, 18(4):389--405.

\bibitem[{Ghosal et~al.(2019)Ghosal, Majumder, Poria, Chhaya, and
  Gelbukh}]{ghosal2019dialoguegcn}
Deepanway Ghosal, Navonil Majumder, Soujanya Poria, Niyati Chhaya, and
  Alexander Gelbukh. 2019.
\newblock Dialoguegcn: A graph convolutional neural network for emotion
  recognition in conversation.
\newblock \emph{arXiv preprint arXiv:1908.11540}.

\bibitem[{Hazarika et~al.(2018{\natexlab{a}})Hazarika, Poria, Mihalcea,
  Cambria, and Zimmermann}]{hazarika2018icon}
Devamanyu Hazarika, Soujanya Poria, Rada Mihalcea, Erik Cambria, and Roger
  Zimmermann. 2018{\natexlab{a}}.
\newblock Icon: Interactive conversational memory network for multimodal
  emotion detection.
\newblock In \emph{Proceedings of the 2018 conference on empirical methods in
  natural language processing}, pages 2594--2604.

\bibitem[{Hazarika et~al.(2018{\natexlab{b}})Hazarika, Poria, Zadeh, Cambria,
  Morency, and Zimmermann}]{hazarika2018conversational}
Devamanyu Hazarika, Soujanya Poria, Amir Zadeh, Erik Cambria, Louis-Philippe
  Morency, and Roger Zimmermann. 2018{\natexlab{b}}.
\newblock Conversational memory network for emotion recognition in dyadic
  dialogue videos.
\newblock In \emph{Proceedings of the conference. Association for Computational
  Linguistics. North American Chapter. Meeting}, volume 2018, page 2122. NIH
  Public Access.

\bibitem[{Hu et~al.(2021)Hu, Liu, Zhao, and Jin}]{hummgcn}
Jingwen Hu, Yuchen Liu, Jinming Zhao, and Qin Jin. 2021.
\newblock Mmgcn: Multimodal fusion via deep graph convolution network for
  emotion recognition in conversation.
\newblock In \emph{Proceedings of the 59th Annual Meeting of the Association
  for Computational Linguistics}.

\bibitem[{Huang et~al.(2017)Huang, Liu, Van Der~Maaten, and
  Weinberger}]{huang2017densely}
Gao Huang, Zhuang Liu, Laurens Van Der~Maaten, and Kilian~Q Weinberger. 2017.
\newblock Densely connected convolutional networks.
\newblock In \emph{Proceedings of the IEEE conference on computer vision and
  pattern recognition}, pages 4700--4708.

\bibitem[{Koval et~al.(2015)Koval, Brose, Pe, Houben, Erbas, Champagne, and
  Kuppens}]{koval2015emotional}
Peter Koval, Annette Brose, Madeline~L Pe, Marlies Houben, Yasemin Erbas,
  Dominique Champagne, and Peter Kuppens. 2015.
\newblock Emotional inertia and external events: The roles of exposure,
  reactivity, and recovery.
\newblock \emph{Emotion}, 15(5):625.

\bibitem[{Li et~al.(2020)Li, Lin, Fu, Si, and Wang}]{li2020hierarchical}
Jiangnan Li, Zheng Lin, Peng Fu, Qingyi Si, and Weiping Wang. 2020.
\newblock A hierarchical transformer with speaker modeling for emotion
  recognition in conversation.
\newblock \emph{arXiv preprint arXiv:2012.14781}.

\bibitem[{Li et~al.(2018)Li, Tao, Schuller, Shan, Jiang, and Jia}]{li2018mec}
Ya~Li, Jianhua Tao, Bj{\"o}rn Schuller, Shiguang Shan, Dongmei Jiang, and Jia
  Jia. 2018.
\newblock Mec 2017: Multimodal emotion recognition challenge.
\newblock In \emph{2018 First Asian Conference on Affective Computing and
  Intelligent Interaction (ACII Asia)}, pages 1--5. IEEE.

\bibitem[{Li et~al.(2017)Li, Su, Shen, Li, Cao, and Niu}]{li2017dailydialog}
Yanran Li, Hui Su, Xiaoyu Shen, Wenjie Li, Ziqiang Cao, and Shuzi Niu. 2017.
\newblock Dailydialog: A manually labelled multi-turn dialogue dataset.
\newblock In \emph{Proceedings of the Eighth International Joint Conference on
  Natural Language Processing (Volume 1: Long Papers)}, pages 986--995.

\bibitem[{Majumder et~al.(2019)Majumder, Poria, Hazarika, Mihalcea, Gelbukh,
  and Cambria}]{majumder2019dialoguernn}
Navonil Majumder, Soujanya Poria, Devamanyu Hazarika, Rada Mihalcea, Alexander
  Gelbukh, and Erik Cambria. 2019.
\newblock Dialoguernn: An attentive rnn for emotion detection in conversations.
\newblock In \emph{Proceedings of the AAAI Conference on Artificial
  Intelligence}, volume~33, pages 6818--6825.

\bibitem[{Mollahosseini et~al.(2017)Mollahosseini, Hasani, and
  Mahoor}]{mollahosseini2017affectnet}
Ali Mollahosseini, Behzad Hasani, and Mohammad~H Mahoor. 2017.
\newblock Affectnet: A database for facial expression, valence, and arousal
  computing in the wild.
\newblock \emph{IEEE Transactions on Affective Computing}, 10(1):18--31.

\bibitem[{Poria et~al.(2017)Poria, Cambria, Hazarika, Majumder, Zadeh, and
  Morency}]{poria2017context}
Soujanya Poria, Erik Cambria, Devamanyu Hazarika, Navonil Majumder, Amir Zadeh,
  and Louis-Philippe Morency. 2017.
\newblock Context-dependent sentiment analysis in user-generated videos.
\newblock In \emph{Proceedings of the 55th annual meeting of the association
  for computational linguistics (volume 1: Long papers)}, pages 873--883.

\bibitem[{Poria et~al.(2019{\natexlab{a}})Poria, Hazarika, Majumder, Naik,
  Cambria, and Mihalcea}]{poria2019meld}
Soujanya Poria, Devamanyu Hazarika, Navonil Majumder, Gautam Naik, Erik
  Cambria, and Rada Mihalcea. 2019{\natexlab{a}}.
\newblock Meld: A multimodal multi-party dataset for emotion recognition in
  conversations.
\newblock In \emph{Proceedings of the 57th Annual Meeting of the Association
  for Computational Linguistics}, pages 527--536.

\bibitem[{Poria et~al.(2019{\natexlab{b}})Poria, Majumder, Mihalcea, and
  Hovy}]{poria2019emotion}
Soujanya Poria, Navonil Majumder, Rada Mihalcea, and Eduard Hovy.
  2019{\natexlab{b}}.
\newblock Emotion recognition in conversation: Research challenges, datasets,
  and recent advances.
\newblock \emph{IEEE Access}, 7:100943--100953.

\bibitem[{Ringeval et~al.(2018)Ringeval, Schuller, Valstar, Cowie, Kaya,
  Schmitt, Amiriparian, Cummins, Lalanne, Michaud et~al.}]{ringeval2018avec}
Fabien Ringeval, Bj{\"o}rn Schuller, Michel Valstar, Roddy Cowie, Heysem Kaya,
  Maximilian Schmitt, Shahin Amiriparian, Nicholas Cummins, Denis Lalanne,
  Adrien Michaud, et~al. 2018.
\newblock Avec 2018 workshop and challenge: Bipolar disorder and cross-cultural
  affect recognition.
\newblock In \emph{Proceedings of the 2018 on audio/visual emotion challenge
  and workshop}, pages 3--13.

\bibitem[{Scherer(2005)}]{scherer2005emotions}
Klaus~R Scherer. 2005.
\newblock What are emotions? and how can they be measured?
\newblock \emph{Social science information}, 44(4):695--729.

\bibitem[{Shen et~al.(2020)Shen, Chen, Quan, and Xie}]{shen2020dialogxl}
Weizhou Shen, Junqing Chen, Xiaojun Quan, and Zhixian Xie. 2020.
\newblock Dialogxl: All-in-one xlnet for multi-party conversation emotion
  recognition.
\newblock \emph{arXiv preprint arXiv:2012.08695}.

\bibitem[{Tao et~al.(2021)Tao, Pan, Das, Qian, Shou, and Li}]{tao2021someone}
Ruijie Tao, Zexu Pan, Rohan~Kumar Das, Xinyuan Qian, Mike~Zheng Shou, and
  Haizhou Li. 2021.
\newblock Is someone speaking? exploring long-term temporal features for
  audio-visual active speaker detection.
\newblock \emph{arXiv preprint arXiv:2107.06592}.

\bibitem[{Vidrascu and Devillers(2005)}]{vidrascu2005annotation}
Laurence Vidrascu and Laurence Devillers. 2005.
\newblock Annotation and detection of blended emotions in real human-human
  dialogs recorded in a call center.
\newblock In \emph{2005 IEEE International Conference on Multimedia and Expo},
  pages 4--pp. IEEE.

\bibitem[{Yang et~al.(2019)Yang, Dai, Yang, Carbonell, Salakhutdinov, and
  Le}]{yang2019xlnet}
Zhilin Yang, Zihang Dai, Yiming Yang, Jaime Carbonell, Russ~R Salakhutdinov,
  and Quoc~V Le. 2019.
\newblock Xlnet: Generalized autoregressive pretraining for language
  understanding.
\newblock \emph{Advances in neural information processing systems}, 32.

\bibitem[{Yu et~al.(2020)Yu, Xu, Meng, Zhu, Ma, Wu, Zou, and Yang}]{yu2020ch}
Wenmeng Yu, Hua Xu, Fanyang Meng, Yilin Zhu, Yixiao Ma, Jiele Wu, Jiyun Zou,
  and Kaicheng Yang. 2020.
\newblock Ch-sims: A chinese multimodal sentiment analysis dataset with
  fine-grained annotation of modality.
\newblock In \emph{Proceedings of the 58th Annual Meeting of the Association
  for Computational Linguistics}, pages 3718--3727.

\bibitem[{Zadeh et~al.(2018)Zadeh, Liang, Poria, Cambria, and
  Morency}]{zadeh2018mosei}
AmirAli~Bagher Zadeh, Paul~Pu Liang, Soujanya Poria, Erik Cambria, and
  Louis-Philippe Morency. 2018.
\newblock Multimodal language analysis in the wild: Cmu-mosei dataset and
  interpretable dynamic fusion graph.
\newblock In \emph{Proceedings of the 56th Annual Meeting of the Association
  for Computational Linguistics (Volume 1: Long Papers)}, pages 2236--2246.

\bibitem[{Zahiri and Choi(2018)}]{zahiri2018emotion}
Sayyed~M Zahiri and Jinho~D Choi. 2018.
\newblock Emotion detection on tv show transcripts with sequence-based
  convolutional neural networks.
\newblock In \emph{Workshops at the thirty-second aaai conference on artificial
  intelligence}.

\bibitem[{Zhao et~al.(2021)Zhao, Li, and Jin}]{zhaomissing}
Jinming Zhao, Ruichen Li, and Qin Jin. 2021.
\newblock Missing modality imagination network for emotion recognition with
  uncertain missing modalities.
\newblock In \emph{Proceedings of the 59th Annual Meeting of the Association
  for Computational Linguistics}, pages 2608--2618.

\end{thebibliography}
\bibliographystyle{acl_natbib}

\clearpage
\appendix
\section{Supplementary}
\label{sec:supplementary}
\subsection{Details of the Active Speaker Detection}
\label{sup:asd}
We observe that the speaker face detection often encounters difficulties in the in-the-wild dialogue scenarios, and the state-of-the-art active speaker detection (ASD) models trained on the English clean dataset normally suffer performance degradation on the Mandarin dataset. 
Therefore, in this work, we propose a two-stage strategy to extract the speaker's faces.
In order to get high-quality faces of the active speakers, we first extract the high-confidence faces of each speaker using a state-of-the-art pre-trained ASD model \cite{tao2021someone}. Then, for the frames that have low detection confidence by the ASD model, we compute the similarity based on the face embeddings\footnote{https://github.com/cydonia999/VGGFace2-pytorch.} between the face in each of these frames and the detected high-confidence speaker's faces, in order to determine which speaker each face in these frames belong to.
The speaker's facial expression information is very important in emotion recognition, and we provide the speaker's facial information even though the detection process is difficult and complicated, while MELD \cite{poria2019meld} did not provide it and did not conduct visual-related experiments.

\subsection{Details of Feature Extractors}
\label{sup:features}
\textbf{Textual Feature Extractor:} We adopt a pre-trained language RoBERTa model in Chinese\footnote{https://huggingface.co/hfl/chinese-roberta-wwm-ext} to extract the word-level textual features. Furthermore, we finetune the pre-trained RoBERTa followed by a classifier on the training set of M$^3$ED to extract more efficient finetuned features. We evaluate utterance-level textual modality performance on the finetuned RoBERTa model. It achieves the weighted-F1 performance of 43.50\% and 45.73\% on the validation and testing sets respectively.

\textbf{Acoustic Feature Extractor:} We adopt a pre-trained speech Wav2Vec model in Chinese\footnote{https://huggingface.co/jonatasgrosman/wav2vec2-large-xlsr-53-chinese-zh-cn} to extract the frame-level acoustic features. Furthermore, we finetune the pre-trained Wav2Vec followed by a classifier on the training set of M$^3$ED to extract more efficient finetuned features. We evaluate utterance-level acoustic modality performance on the finetuned Wav2Vec model. It achieves the weighted-F1 performance of 48.56\% and 45.92\% on the validation and testing sets respectively.

\textbf{Visual Feature Extractor:} We adopt a pre-trained facial expression recognition DenseNet model to extract the face-level visual features, which is trained on the combination of the FER+ and AffectNet two benchmark corpus (Tabel.~\ref{tab:facialexp} ). It achieves the Weighted-accuracy and F1-score performance of 63.54\% and 52.94\% on the combined validation set respectively. 

\begin{table}[t]
\caption{The distribution of facial expression recognition datasets. $neu$, $hap$, $sur$, $ang$, $dis$ and $con$ denotes neutral, happy, surprise, anger, disgust and contempt respectively.}
\scalebox{0.8}{
\begin{tabular}{c|cc|cc|cc}
\hline
  & \multicolumn{2}{c|}{FER+}  & \multicolumn{2}{c|}{AffectNet} & \multicolumn{2}{c}{Total}  \\ 
          & train    & val   & train     & val    & train & val  \\
\hline
neu   & 10,342 & 1,342  &  74,874  &  500  &   85,216   &  1842  \\
hap   & 7,526 &  898  &  134,415 &  500  &  141,941  &  1398 \\
sur   & 3,576 &  458  &  14,090  &  500  &  17,666   &  958  \\
sad   & 3,530 &  416  &  25,459  &  500  &   28,989   &  916   \\
ang   & 2,464 &  319  &  24,882  &  500  & 27,346   &  819    \\
dis   & 654 &  36  &  3,803   &  500  & 3,996    &  536  \\
fear  & 193 &  75  &  6,378   &  500  & 7,032   &  575     \\
con   & 167 &  25   &  3,750   &  499  &  3,917   &  524  \\
\hline
\end{tabular}
\label{tab:facialexp}
}
\end{table}

\subsection{Extra Experimental Results Analysis.}
\begin{figure}[h]
    \centering
     \begin{minipage}{0.49\linewidth}
      \centerline{\includegraphics[width=3.5cm]{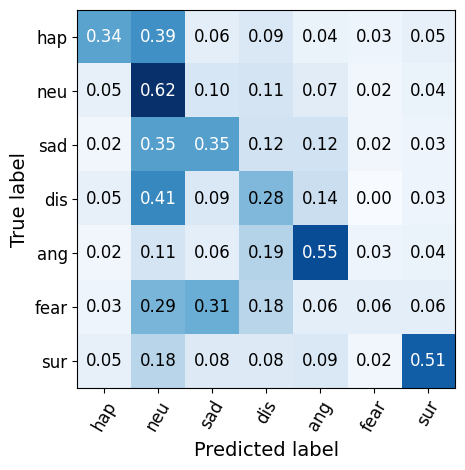}}
      \centerline{\small{(b) DialogRNN}}
    \end{minipage}
    \begin{minipage}{0.49\linewidth}
      \centerline{\includegraphics[width=3.5cm]{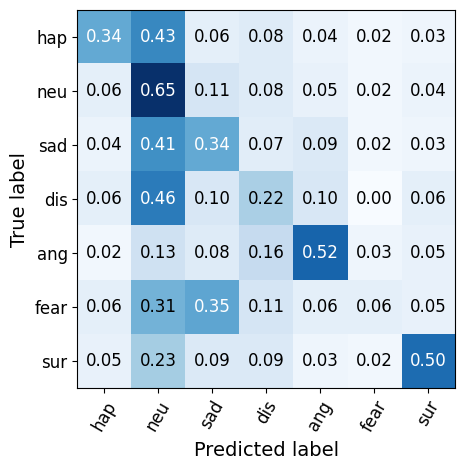}}
      \centerline{\small{(b) Ours MDI}}
    \end{minipage}
    \caption{Confusion matrices of DialogueRNN and MDI models under the $\{l,a,v\}$ multimodal condition.}
    \label{fig:confusion}
\end{figure}

Figure.~\ref{fig:confusion} presents the confusion matrices of DialogueRNN and our MDI dialogue-level models under the $\{l,a,v\}$ multimodal condition. Both models perform badly for recognizing the fear emotion, which relates to the limited number of training instances for the fear emotion. It demonstrates the class imbalance issue is a challenging problem for both models. We also observe a high confusing rate between sad, anger, and disgust emotion categories since these emotions are more likely to occur at the same time (the top 5 blended emotions indeed come from these 3 categories), which makes them more difficult to disambiguate. In the future, we will explore effective solutions to deal with the emotion imbalance challenge and learn multi-label emotion classification.

\end{document}